# Temporal Fusion Transformer for Multi-Horizon Probabilistic Forecasting of Weekly Retail Sales

Santhi Bharath Punati
*Sr. Product Manager*
Sunbelt Rentals
Fort Mill, South Carolina, USA
0009-0004-5252-0611

Sandeep Kanta
*Researcher*
Northeastern University
Dallas, Texas, USA
0009-0000-7518-1115

Udaya Bhasker Cheerala
*Senior Consultant*
Independent Researcher
Bay Area, California, USA
0009-0000-9338-3327

Madhusudan G Lanjewar
*Research Technician*
School of Physical and Applied Sciences
Goa University
Goa, India
0000-0002-9670-3020

Praveen Damacharla
*Research Scientist*
KINETICAI INC
The Woodlands, Texas, USA
praveen@kineticai.com
0000-0001-8058-7072

*Abstract*—Accurate multi-horizon retail forecasts are critical for inventory and promotions. We present a novel study of weekly Walmart sales (45 stores, 2010–2012) using a Temporal Fusion Transformer (TFT) that fuses static store identifiers with time-varying exogenous signals (holidays, CPI, fuel price, temperature). The pipeline produces 1–5-week-ahead probabilistic forecasts via *QuantileLoss*, yielding calibrated 90% prediction intervals and interpretability through variable-selection networks, static enrichment, and temporal attention. On a fixed 2012 hold-out dataset, TFT achieves an RMSE of $ 57.9k USD per store-week and an $R^2$ of 0.9875. Across 5-fold chronological cross-validation, the averages are RMSE = $ 64.6k USD and $R^2$ = 0.9844, outperforming XGB, CNN, LSTM, and CNN-LSTM baseline models. These results demonstrate practical value for inventory planning and holiday-period optimization, while maintaining model transparency.

## I. INTRODUCTION

The rapid growth of e-commerce impacts communities, processing hundreds of thousands of trades daily and generating massive, complex, and high-dimensional data [1]. This creates unprecedented potential for data-driven, factual sales forecasting. Companies rely heavily on it to optimize products, allocate resources, and plan strategically. Conventional forecasting models relied solely on a company's sales data, but the rise of big data and online transactions has transformed this approach [1]. Generally, these datasets contain high noise, variability, nonlinear behavior, missing values, and disparities [2], [3]. Similarly, consumer demands, socioeconomic trends, market competition, seasonality, and unforeseen natural disasters create unpredictable sales patterns that challenge traditional methods.

Walmart, one of the most prominent retail companies in the United States with operations across multiple regions, is significantly influenced by external factors such as economic trends and seasonal festivals. The sales forecasting becomes challenging due to unexpected patterns rendered by promotional cost declines and seasonal vacations like Christmas, which significantly affect sales. Traditional forecasting methods, often based on time series models and linear regression, struggle to handle such vast and complex datasets due to their inability to capture nonlinear relationships and interactions among several variables [1]. To address these challenges, machine learning (ML) and deep learning (DL) can handle massive amounts of data, support large-scale computation, and model complex, nonlinear patterns [4]. DL approaches such as Convolutional Neural Networks (CNN) and Long Short-Term Memory (LSTM) networks have shown significant efficacy in sales forecasting [3], [5], [6]. ML/DL excels in automated feature extraction, temporal relationship modeling, and addressing complex customer and market trends.

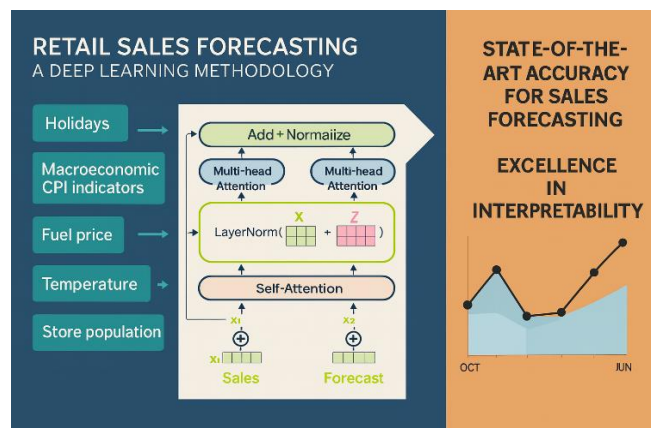

Fig. 1: A Generic TFT Retail Sales Forecasting Framework

Despite their strong performance, these models rely on historical statistics and auxiliary characteristics, often overlooking the periodic (seasonal) and non-periodic (irregular) variations in sales. This limits their generalizability and ability to adjust to real-world dynamics. In such cases, the Transformer design, which is built on a system of encoders and decoders and powered by self-attention processes, is particularly good at simulating long-range relationships inside sequences [7]. Its ability to handle complicated temporal correlations by capturing intricate temporal dynamics underlying demand fluctuation has resulted in extensive use in various domains [1].

Studies such as [1], [8] applied Transformers to the Corporación Favorita dataset, reporting RMSLE values of 0.51 and 0.54, respectively, but with limited treatment of complex seasonality. In contrast, [9] proposed Aliformer, a bidirectional Transformer with self-attention and a forward-looking training approach, designed to capture both historical trends and contextual factors, and obtained MSE and MAE of 0.154 and 0.229, respectively. However, training and hyperparameter adjustment were resource-intensive. In [10] Autoformer demonstrated strong performance across six benchmark datasets (e.g., ETT: MSE = 0.065, MAE = 0.189), but required substantial processing resources. In [11], FEDTransformer integrated Transformers with frequency-enhanced seasonal-trend decomposition to improve global structure learning while reducing computational cost, achieving MSE and MAE of 0.287 and 0.276 on the ETT and Exchange datasets, respectively. In [12], applying Transformers to Walmart sales yielded MSE = 25.76, MAE = 3.12, and $R^2 \approx 0.95$, though performance was sensitive to market volatility.

Existing studies often employed those inadequately included static variables, limiting their ability to leverage fixed yet crucial features such as location, category, or product type, which can substantially influence forecasts. Furthermore, most previous techniques only provided point forecasts without quantile outputs, rendering it challenging to quantify uncertainty in forecasting or estimate risk under changing conditions. This shortcoming reduces their applicability in decision-making procedures requiring probabilistic insights. In contrast, the TFT addresses these issues by natively including static variables into the structure and generating multi-horizon quantile predictions. Consequently, TFT produces richer and more interpretable predictions, incorporating trends and uncertainty estimates, and thus offers a more robust and informative alternative to prior models.

Accurate sales forecasting remains challenging due to complex, nonlinear patterns impacted by variables such as promotions and seasonality. Traditional models, sometimes, do not capture important cyclical and non-cyclical fluctuations. To address this, the present work in Fig. 1 employs a Transformer with attention mechanisms to forecast weekly Walmart sales data. The model captures seasonal transformations and external economic forces impacting retail performance, thereby reducing operational uncertainty, enhancing inventory planning, cutting costs, and boosting economic performance. The primary contributions of this work are as follows:

- Development of a unified TFT-based forecasting pipeline that incorporates external factors and static store-level characteristics.
- Provision of accurate and interpretable uncertainty estimates through attention-based mechanisms.
- Demonstration of model robustness via 5-fold cross-validation and detailed baseline assessments.

## II. MATERIAL AND METHOD

Fig. 2(a) depicts a pipeline of the suggested architecture for anticipating Walmart sales data. Fig. 2(b) shows the typical TFT architecture [13]. This study used the Walmart sales dataset, which was then pre-processed to ensure data reliability and uniformity. The pre-processed data were then fed into four forecasting models – Extreme Gradient Boosting (XGB), 1D Convolutional Neural Network (CNN), Long Short-Term Memory (LSTM), and a hybrid CNN-LSTM – along with TFT. Model performance was evaluated and compared to identify the most effective sales forecasting approach.

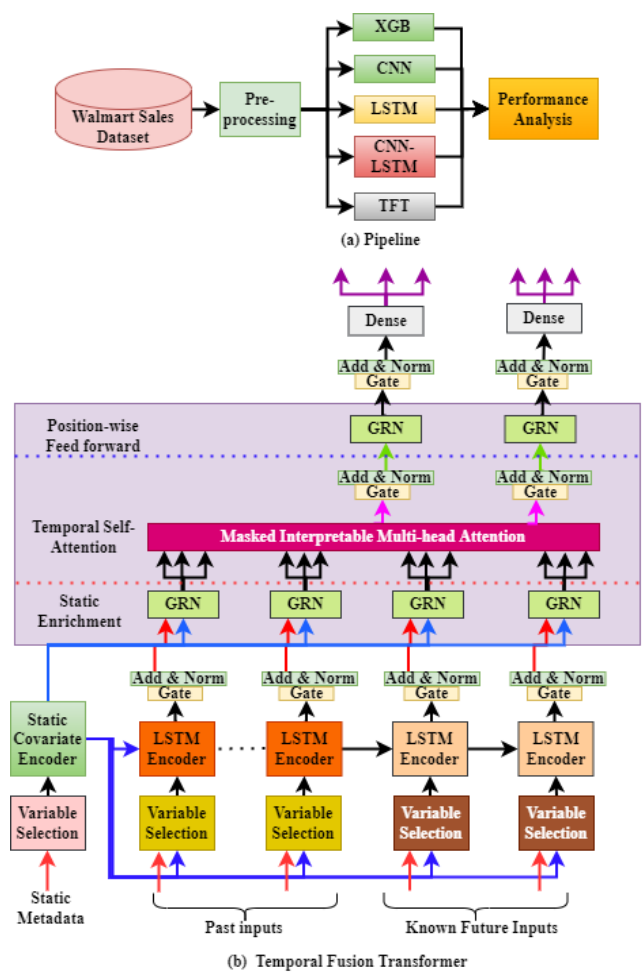

Fig. 2: Proposed framework for sales forecasting using the TFT: (a) end-to-end pipeline of the presented methodology, and (b) internal TFT block architecture incorporating static covariate encoder, GRN, and multi-head

### A. Dataset and Pre-processing

In [14], an online dataset on Kaggle containing historical weekly sales information from February 2010 to November 2012, across forty-five Walmart retail outlets, was used. Walmart seeks to increase its sales through effective forecasting systems that minimize stockouts and optimize inventory planning. Sales variability is highly impacted by temporal events, which cause nonlinear oscillations in customer purchasing behavior. At the store level, the dataset comprises many variables: a unique store identifier, sales week, the objective variable (*Weekly_Sales*), a binary holiday indication (*Holiday_Flag*), ambient temperature, gasoline price, Consumer Price Index (CPI), and the unemployment (UEMP) rate. Table 1 displays the illustrative statistics for the Walmart weekly sales dataset. The variable Store varies from 1 to 45, with a mean of 23 and a standard deviation (SD) of about 12.9, showing an even distribution of data across all shop locations.

The target variable *Weekly_Sales* (Sales) has a mean value of 1,046,964.8 with an SD of 564,366.6, ranging from 209,986.2 to 3818686.5, reflecting substantial temporal and geographical variability. The *Holiday_Flag* (Holiday) variable has a mean value of 0.07, suggesting only 7% of the observations are linked to vacation duration. Temperature (Temp) readings vary from -2.06°F to 100.14°F, with a mean of 60.66°F and an SD of 18.44°F, indicating data coverage of a wide variety of climate conditions. *Fuel_Price* (FP) ranges from $2.47 to $4.47, with a mean of $3.36 and an SD of 0.46. The CPI runs from 126.06 to 227.23, with a mean of 171.58, illustrating inflationary trends over time. UEMP rates range from 3.88%-14.31%, with an average of around 8.00% reflecting a broad range of economic circumstances.

Fig. 3 depicts weekly sales trends and a correlation heatmap. The heatmap reveals poor pairwise linear correlations, reinforcing the need for nonlinear models like TFT for effective sales forecasting. The dataset was pre-processed to organize the time series for effective modeling, including addition of a new column (*t_idx*) to index weeks from the earliest date. Data were then chronologically sorted by Store and Date to preserve temporal order. The Store column was converted into a string and managed as a category variable. Finally, *Weekly_Sales* was log-transformed to normalize its distribution and the effect of outliers.

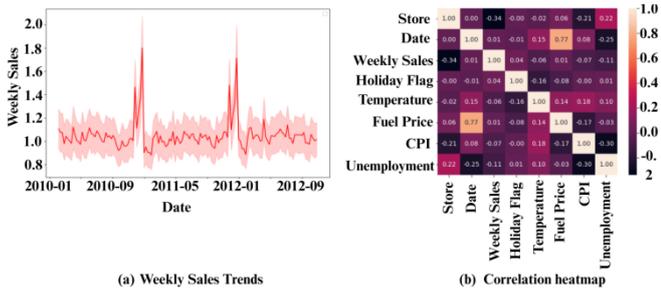

Fig. 3: Weekly sales trends across Walmart stores and a correlation heatmap of important characteristics, revealing significant negative correlation of *Weekly_Sales* with Store ID, a strong Date-Fuel Price connection, and a moderate negative CPI-Unemployment correlation.

It also assists in reducing the skewness produced by large numbers, and prevents them from controlling model training while enhancing forecast stability and reliability. In addition, it enables the model to detect more relevant changes instead of being unduly sensitive to huge absolute values.

TABLE I. DESCRIPTIVE STATISTICS FOR THE WALMART WEEKLY SALES DATASET

|  | Store | Sales | Holiday | Temp | FP | CPI | UEMP |
|---|---|---|---|---|---|---|---|
| **Count** | 6435 | 6435 | 6435 | 6435 | 6435 | 6435 | 6435 |
| **Mean** | 23 | 1046964.8 | 0.07 | 60.66 | 3.36 | 171.58 | 8.00 |
| **SD** | 12.9 | 564366.6 | 0.26 | 18.44 | 0.46 | 39.36 | 1.88 |
| **Min** | 1 | 209986.2 | 0.00 | -2.06 | 2.47 | 126.06 | 3.88 |
| **25%** | 12 | 553350.1 | 0.00 | 47.46 | 2.93 | 131.74 | 6.89 |
| **50%** | 23 | 960746.0 | 0.00 | 62.67 | 3.44 | 182.62 | 7.87 |
| **75%** | 34 | 1420158.7 | 0.00 | 74.94 | 3.73 | 212.74 | 8.62 |
| **Max** | 45 | 3818686.5 | 1.00 | 100.14 | 4.47 | 227.23 | 14.31 |

### B. Temporal Fusion Transformer (TFT)

The TFT is an advanced DL structure, which integrates recurrent layers, attention mechanisms, and gating structures [13]. It captures relevant static and temporal characteristics at each time step to improve forecasting performance. Static covariates (Store ID) are incorporated using a static covariate encoder to provide global context across sequences, while LSTM layers paired with multi-head attention capture both short- and long-term temporal patterns. Gated Residual Networks (GRNs) further model complex correlations while maintaining model stability. TFT successfully considers exogenous variables such as CPI, fuel price, and temperature, resulting in accurate sales forecasting.

The TFT was set up with an encoder length of 52 weeks and a prediction length of 5 weeks, allowing it to learn from one year of history data and forecast five weeks. A training cut-off assured temporal causality by removing future data during the training phase. The *TimeSeriesDataSet* object organized the data utilizing *Weekly_Sales* as the objective, and it comprises static category characteristics (e.g., Store ID), known real-valued variables (e.g., CPI, Temp), and unknown time-varying variables. The hyperparameters were tuned with grid search for the optimum outcome. The authors found the best hyperparameters of batch size 16, *QuantileLoss* (0.1, 0.5, and 0.9), a learning rate of 0.01, a hidden size of 64, an attention head size of 4, and a dropout rate of 0.2.

### III. RESULTS

An HP laptop with a Ryzen 5600H processor and 16GB of RAM was used with Python version 3.9. The dataset was split into training and validation sets at an 80:20 ratio, and 5-fold cross-validation was used to cross-validate the model. The TFT training time was approximately 780 seconds. Dataset variables were normalized using the *StandardScaler* technique to achieve zero mean and unit variance. Conversely, the *MinMaxScaler* was used to scale the target variable's values within a predefined range, allowing for more stable model training and improved convergence. Model performance was evaluated using regression metrics such as $R^2$, RMSE, MAE, and SMAPE [12].

### A. TFT Performance on Walmart Dataset

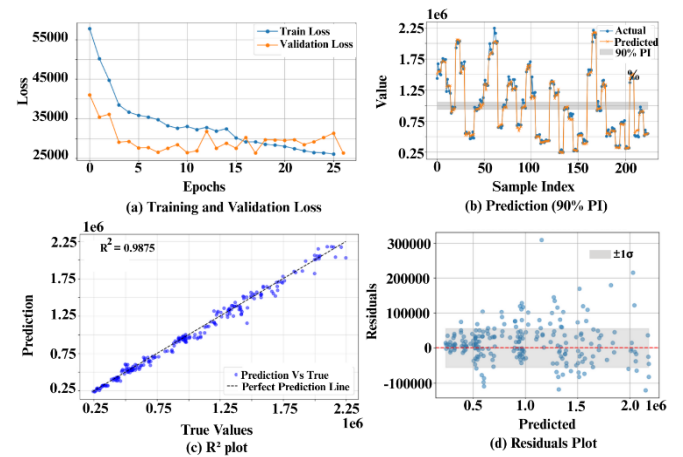

Fig. 4: Performance of the TFT in sales forecasting using the Walmart dataset was examined using an 80/20 train-test split and a 5-week forecast horizon at the weekly interval level.

The original dataset contained eight columns, which expanded to ten after pre-processing. The data were split into training (5,228 x 10) and validation (1,307 x 10) sets, and subsequently used to train the TFT. Fig. 4(a) illustrates that training (2.63e+4) and validation (3.33e+4) losses fall steadily across epochs before stabilizing, showing effective learning without overfitting. Fig. 4(b) shows the expected sales numbers and 90% PI ranges, indicating high forecasting capacity with defined uncertainty. In Fig. 4(c), the predicted and real values are firmly packed around the ideal diagonal line, indicating remarkable predictive accuracy with an R² value of 0.9875. Fig. 4(d) displays residuals randomly distributed around zero, confirming that the model captures underlying data structures and that errors are random.

Fig. 5 depicts the efficacy and interpretability of TFT in weekly sales forecasting. Fig. 5(a) illustrates the attention distribution across prior time steps, indicating the historical periods most influential in shaping predictions. Fig. 5(b) shows store-level predictions that adequately capture changes across various stores, indicating the model's resilience in dealing with cross-sectional variances and temporal dynamics.

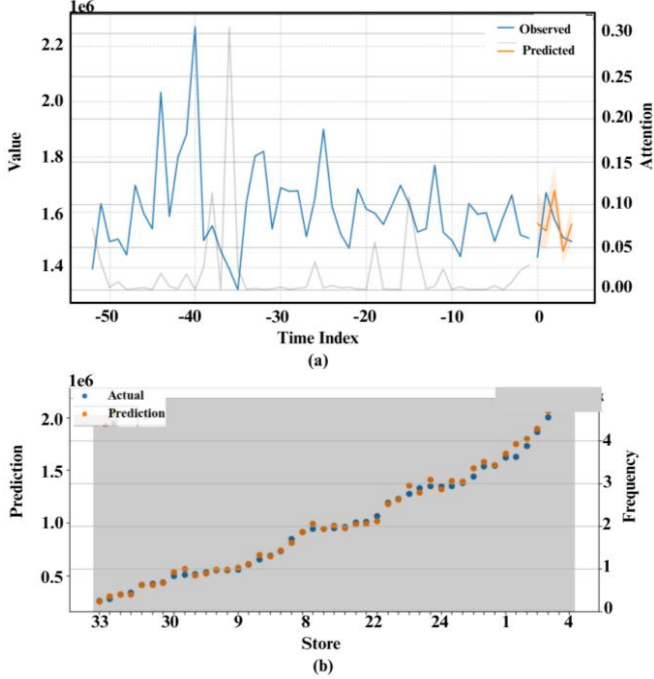

Fig. 5: (a) Attention weights across time lags, and (b) ranked prediction frequencies for different stores.

*B. 5-fold Cross Validation Performance*

Table II displays the model's effectiveness throughout 5-fold cross-validation utilizing RMSE, MAE, $R^2$, and SMAPE. The TFT achieved the ranges for RMSE (57973.84 to 70750.64), MAE (41504.23-47360.05), $R^2$ (0.9814-0.9875), and SMAPE (4.51-5.00%). The Fold 2 performed the best among the folds, suggesting minimal error and higher forecasting effectiveness. The model's 5-fold average (Ave) RMSE = 64626.95 ± 4674.19, MAE = 44505.43 ± 2244.07, $R^2$ = 0.9844 ± 0.01, and SMAPE = 4.74 ± 0.19% indicated good and consistent prediction accuracy at all folds.

TABLE II. PERFORMANCE OF TFT WITH 5-FOLD CROSS-VALIDATION ACROSS DISTINCT DATA SPLITS

| Folds | RMSE | MAE | $R^2$ | SMAPE |
|---|---|---|---|---|
| 1 | 66665.43 | 45735.06 | 0.9835 | 4.85% |
| 2 | 57973.84 | 41504.23 | 0.9875 | 4.51% |
| 3 | 64281.58 | 43297.28 | 0.9846 | 4.62% |
| 4 | 63463.26 | 44630.54 | 0.9850 | 4.73% |
| 5 | 70750.64 | 47360.05 | 0.9814 | 5.00% |
| Ave | 64626.95 ± 4674.19 | 44505.43 ± 2244.07 | 0.9844 ± 0.01 | 4.74±0.19% |

*C. Comparative Analysis*

Table III compares the effectiveness of XGB, CNN, LSTM, CNN-LSTM, and TFT. The hyperparameters of all models were: XGB: colsample_bytree = 0.8, learning_rate = 0.05, max_depth = 7, min_child_weight = 4, n_estimators = 30, and subsample = 0.8; CNN-1D: 32 filters (kernel size = 4), ReLU activation, and same padding; LSTM: 50 units; CNN-LSTM: combine CNN1D and LSTM. Common settings (CNN, LSTM, and CNN-LSTM) were set to: one-unit dense layer, MSE loss, Adam optimizer, and learning rate = 0.001.

From Table III, TFT surpasses all other models for all criteria, demonstrating higher accuracy in forecasting. CNN-LSTM ranks second, with RMSE, MAE, $R^2$, and SMAPE of 117231.56, 87954.85, 0.9573, and 11.95%, respectively. XGB, CNN, and LSTM have lower efficacy and higher error rates. Fig. 6 further verifies the TFT's efficacy compared to the XGB, CNN, LSTM, and CNN-LSTM.

TABLE III. COMPARATIVE ANALYSIS OF XGB, CNN, LSTM, CNN-LSTM, AND TFT

| Models | RMSE | MAE | $R^2$ | SMAPE |
|---|---|---|---|---|
| XGB | 150573.92 | 118947.20 | 0.9296 | 14.65% |
| CNN | 187195.28 | 135757.27 | 0.8912 | 18.27% |
| LSTM | 137666.79 | 101496.75 | 0.9412 | 16.99% |
| CNN-LSTM | 117231.56 | 87954.85 | 0.9573 | 11.95% |
| TFT | 57973.84 | 41504.23 | 0.9875 | 4.51% |

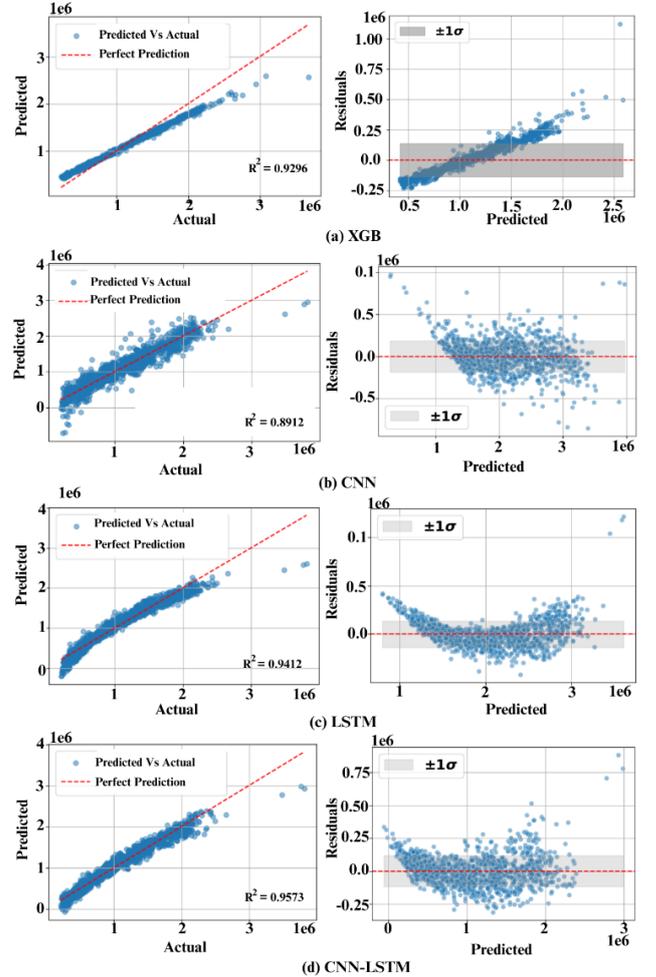

Fig. 6: $R^2$ plot and Residual plot of (a) XGB, (b) CNN, (c) LSTM, and (d) CNN-LSTM.

The TFT scatter plots in Fig. 4(c) and 4(d) display forecasts closely aligned with actual values with minimum variance, while the residuals are densely packed around zero, suggesting reduced systematic bias. In contrast, XGB demonstrates strong alignment for larger values but exhibits significant residual dispersion at smaller scales. CNN, LSTM, and CNN-LSTM display greater scatter around the ideal forecast line and show distinct residual patterns, implying underfitting and insufficient capture of all temporal correlations.

## IV. Discussion

Weekly sales forecasting using the Walmart dataset is challenging. However, TFT provides an effective approach for multivariate time series data incorporating both temporal and static factors. The 5-week prediction horizon yields meaningful forward-looking insights for operational planning, including inventory management and workforce allocation. Encoding Store as a static category feature allows the model to learn store-specific behaviors via embeddings, thereby enhancing customization and forecast accuracy across retail locations. Moreover, the *QuantileLoss* function was employed to calculate prediction intervals, a critical feature for retail demand forecasting where over- or under-estimation may lead to stockouts or excess inventory holding costs. Multi-head attention mechanisms enhance interpretability by emphasizing key time steps and variables, whereas networks stabilize training and capture nonlinear relations. The model's configuration also includes batching and learning rate scheduling to improve training efficiency and scalability for large datasets.

Existing ML/DL methods, including Transformers, have improved forecasting performance by emphasizing context-aware attention. However, they often neglect the integration of static variables and uncertainty quantification, require extensive design tuning and provide limited support for uncertainty estimations. In contrast, the TFT provides a comprehensive understanding of the forecasting problem – an essential requirement in retail settings such as Walmart, where stocking decisions depend on both anticipated demand and its variability. Authors in [12] achieved high $R^2$ and low error metrics employing standard attention with structured data. This work employs TFT as a more principled framework to integrate macro-economic indicators, environmental data, and temporal context, thus enhancing robustness under diverse real-world conditions. While [1] reduced errors by explicitly modelling seasonal and promotional effects, the TFT's dynamic variable selection network automates this process, increasing scalability and adaptability to new data sources and areas.

## V. Conclusion

The TFT model successfully forecasted weekly sales from the Walmart dataset. Unlike traditional models and other Transformer variants, TFT combines static covariates and dynamic time-varying features such as holidays, CPI, fuel price, and temperature, permitting the framework to acquire short- and long-term dependencies while offering probabilistic forecasts. The suggested model accurately forecasted weekly sales with high predictive performance (RMSE = 57973.84, MAE = 41504.23, $R^2$ = 0.9875, and SMAPE = 4.51%). In the future, we want to verify the model across numerous benchmark datasets to examine its generalizability and resilience in various forecasting situations.